\documentclass[letterpaper, 10 pt, conference]{ieeeconf}  

\usepackage{FG2023}

\FGfinalcopy 

\IEEEoverridecommandlockouts                              
\usepackage{graphics} 
\usepackage{epsfig} 
\usepackage{mathptmx} 
\usepackage{times} 
\usepackage{amsmath} 
\usepackage{amssymb}  
\usepackage{adjustbox}
\usepackage{makecell}
\usepackage{booktabs}

\usepackage{hyperref}
\usepackage{makebox}

\def\ie{\emph{i.e.}}
\def\etal{\emph{et al.}}

\def\eg{\emph{e.g.}}

\definecolor{cfgreen}{rgb}{0.0, 0.42, 0.24}

\newcommand{\datasetname}{Florence}

\def\FGPaperID{****} 

\title{\LARGE \bf
The \datasetname{} 4D Facial Expression Dataset}

\author{\parbox{16cm}{\centering
    {\large Filippo Principi$^1$, Stefano Berretti$^1$, Claudio Ferrari$^2$,  Naima Otberdout$^3$, \\Mohamed Daoudi$^4$, Alberto Del Bimbo$^1$}\\
    {\normalsize
    $^1$ Media Integration and Communication Center, University of Florence, Italy\\
    $^2$ Department of Engineering and Architecture, University of Parma, Italy\\
    $^3$ Ai movement - University Mohammed IV Polytechnic, Rabat, Morocco\\ 
    $^4$ Univ. Lille, CNRS, Centrale Lille, Institut Mines-Télécom, UMR 9189 CRIStAL, F-59000 Lille, France\\
    IMT Nord Europe, Institut Mines-Télécom, Univ. Lille, Centre for Digital Systems, F-59000 Lille, France\\
}
    }
}

\begin{document}

\ifFGfinal
\thispagestyle{empty}
\pagestyle{empty}
\else
\author{Anonymous FG2023 submission\\ Paper ID 89 \\}
\pagestyle{plain}
\fi
\maketitle

\begin{abstract}
Human facial expressions change dynamically, so their recognition / analysis should be conducted by accounting for the temporal evolution of face deformations either in 2D or 3D. While abundant 2D video data do exist, this is not the case in 3D, where few 3D dynamic (4D) datasets were released for public use. The negative consequence of this scarcity of data is amplified by current deep learning based-methods for facial expression analysis that require large quantities of variegate samples to be effectively trained. With the aim of smoothing such limitations, in this paper we propose a large dataset, named \datasetname{} 4D, composed of dynamic sequences of 3D face models, where a combination of synthetic and real identities exhibit an unprecedented variety of 4D facial expressions, with variations that include the classical neutral-apex transition, but generalize to expression-to-expression. 
All these characteristics are not exposed by any of the existing 4D datasets and they cannot even be obtained by combining more than one dataset. We strongly believe that making such a data corpora publicly available to the community will allow designing and experimenting new applications that were not possible to investigate till now. To show at some extent the difficulty of our data in terms of different identities and varying expressions, we also report a baseline experimentation on the proposed dataset that can be used as baseline. 
\end{abstract}

\section{INTRODUCTION}
Facial expressions play a primary role in interpersonal relations and are one fundamental way to convey our emotional state~\cite{Dang2021icact}. 
The automatic analysis of facial expressions focused first on images and videos, with rare examples using 3D data~\cite{ariano2021action, ferrari2015dictionary, sandbach2012static, zhu2019discriminative}. These initial studies focused more on datasets with posed expressions~\cite{Lucey2010cvprw}, impersonated by actors. The trend is now moving towards spontaneous (not posed) datasets, with some examples of in-the-wild acquisitions~\cite{Dhall2011iccvw, Kossaifi2017imavis}. 
For a summary description of 2D datasets for macro and micro facial expression analysis, we refer to the survey in~\cite{Guerdelli2022sensors}.

Thanks to the rise of powerful deep learning based solutions, the interest in applying expression recognition/generation on 3D and 4D data is growing rapidly. However, with such paradigm, the variety of applications that can be designed and their effectiveness are mainly bounded by the volume and variety of available data to train the models. Thus, it is evident the importance of collecting sufficiently large and variegate datasets, which need to be designed for each specific task. The greater cost and challenges in acquiring dynamic 3D (4D) data is at the base of their limited availability. 3D data are also intrinsically more difficult to process. For example, reducing 3D scans to a same number of vertices connected by a same topology is by itself a task which is complicated to solve in general, while a dense correspondence between scans is, in the majority of cases, a prerequisite to make learning methods to work properly. In addition to being difficult to acquire, 3D data are also difficult to annotate automatically, while manual intervention is impractical and error prone for large volumes of data. A possible workaround is constituted by the creation of synthetic data. For example, this direction has been successfully taken in the case of human body, with a relevant example given by the (not public) dataset used for training the 3D position detector of the skeleton joints in the Kinect for Xbox 360~\cite{Shotton2011cvpr}. Also, if properly designed, the domain gap between real and synthetic 3D data can be made significantly small~\cite{wood2021iccv}, reducing the performance drop that is usually observed in these cases.  

In this paper, we propose the ``\datasetname{} 4D Facial Expression Dataset'' (\datasetname{} 4D for short), the first collection of 3D dynamic sequences of facial expressions that includes many identities and long sequences with composed expressions.
The peculiar characteristics of our dataset are:
\begin{itemize}
    \item a large number of identities, both real and synthetic, is included, mostly balanced between male and female;
    \item all the 3D models share the same number of vertices and connectivity among them, \ie, same topology:
    \item large number of expressions (70 expressions in total), most of them being variations or combinations of 8 prototypical ones, so to cover a much larger spectrum;
    \item expression sequences with a large variability: sequences with neutral to apex transitions, as well as with transitions between two apex expressions are included; 
    \item the temporal evolution of the sequences is generated with randomized velocity for improved realism.
\end{itemize}

\begin{table*}[!ht]
\centering
\caption{Most popular HR 3D/4D facial expression datasets. \datasetname{} 4D Facial Expressions dataset is unique in providing both real and synthetic 3D models, with an unprecedented number of sequences, also including expression-to-expression transitions}
\begin{tabular}{p{0.14\linewidth}cp{0.2\linewidth}p{0.4\linewidth}p{0.1\linewidth}}
\hline
\textbf{Datasets} & \textbf{\#IDs} & \textbf{Pose} & \textbf{Expressions} & \textbf{\# sequences} \\
\hline
FRGC v2.0~\cite{phillips:2006} & 466 & slight & small cooperative expressions--\textit{disgust}, \textit{happiness}, \textit{sadness}, \textit{surprise},--and large uncooperative expression, not categorized & static \\
Bosphorus~\cite{savran2008bosphorus} & 105 & 13 yaw and pitch systematic head rotations) & neutral plus 6 basis expressions, selected AUs & static \\
BU-3DFE~\cite{yin20063d} & 100 & frontal & neutral plus 6 basis expressions & static \\
Florence 2D/3D~\cite{Bagdanov:2011} & 61 & 2 frontal plus 2 side & neutral & static \\
\hline
BU-4DFE~\cite{Yin2008fg} & 101 & frontal & neutral plus 6 basis expressions & 606 \\
CoMA~\cite{ranjan2018generating} & 12 & frontal & 12 extreme and asymmetric expressions & 144 \\
D3DFACS~\cite{cosker2011facs} & 10 & frontal & Action Units & 519 \\
VOCASET~\cite{VOCA2019} & 12 & frontal & Speech & 480 \\
\hline
\textbf{\datasetname{} 4D (Ours)} & 95 & frontal & 70 expressions as variations/combinations of 8 categories: \textit{anger}, \textit{fear}, \textit{sadness}, \textit{disgust}, \textit{surprise}, \textit{anticipation}, \textit{trust}, \textit{joy} & 205,200 \\
\hline
\end{tabular}
\label{tab:datasets}
\end{table*}

\noindent
All the above characteristics cannot be found in currently available datasets, and can likely open the way to the exploration of completely new tasks. To show that the synthetic data we have generated can effectively complement real data, we compare it with other common benchmarks of real scans in the task of landmark-based 3D face fitting. 

\textcolor{black}{The dataset is available at the following link} \url{www.micc.unifi.it/resources/datasets/florence-4d-facial-expression/}


\section{RELATED WORK}\label{sect:related-work}
In the following. we summarize the existing 3D dynamic datasets for facial expression analysis. We also refer to the 3D static datasets because they can be used in combination with the dynamic ones in some specific application. We restrict our analysis to high-resolution (HR) 3D datasets, while there are also some low resolution datasets either static or dynamic that have been captured with low-cost, low-resolution cameras like Kinect. Notable examples are the Eurecom dataset~\cite{eurecom}, the IIIT-D RGB-D face database~\cite{iiitd}, and FaceWarehouse~\cite{cao2013facewarehouse}. 
Table~\ref{tab:datasets} compares the main characteristics of the HR datasets discussed below.

\paragraph{Static 3D datasets}
The Face Recognition Grand Challenge dataset (FRGC v2.0)~\cite{phillips:2005, phillips:2006} includes 3D face scans partitioned in three sets, namely, \textit{Spring2003} (943 scans of 277 individuals), \textit{Fall2003}, and \textit{Spring2004} (4,007 scans of 466 subjects in total).  
Individuals have been acquired with frontal view from the shoulder level, with very small pose variations. About 60\% of the faces have neutral expression, and the others show expressions of \textit{disgust}, \textit{happiness}, \textit{sadness}, and \textit{surprise}. 
The Bosphorus dataset~\cite{savran2008bosphorus} comprises $4,666$ high-resolution scans of $105$ individuals. There are up to $54$ scans per subject, which include prototypical expressions and facial Action Unit (AU) activation. The raw scans of Bosphorus have an average of $30K$ vertices on the face region. This dataset also contains rotated and occluded scans. 
The Bimghamton University 3D facial expression dataset~\cite{yin20063d} comprises $100$ subjects (56 females and 44 males) in the six prototypical expressions (angry, disgust, fear, happy, sad, surprise) plus neutral, each expression being reproduced at four intensity levels, from low to exaggerated. Therefore, there are $25$ 3D expression scans for each subject, resulting in $2,500$ 3D facial expression scans in total. The subjects vary in gender, ethnicity (White, Black, East-Asian, Middle-east Asian, Indian, and Hispanic Latino) and age (from 18 to 70).  
The Florence 2D/3D hybrid face dataset~\cite{Bagdanov:2011} includes 3D face scans and video acquisitions of 61 subjects. For each subject, there are two neutral and two side scans plus RGB video recordings both posed indoor and unconstrained outdoor. Being one of the few datasets including both 2D videos and 3D models of same subjects, it has found large use in evaluating methods that reconstruct the 3D geometry of the face from 2D data. 

\begin{figure}[!t]
\centering
\includegraphics[width=\linewidth]{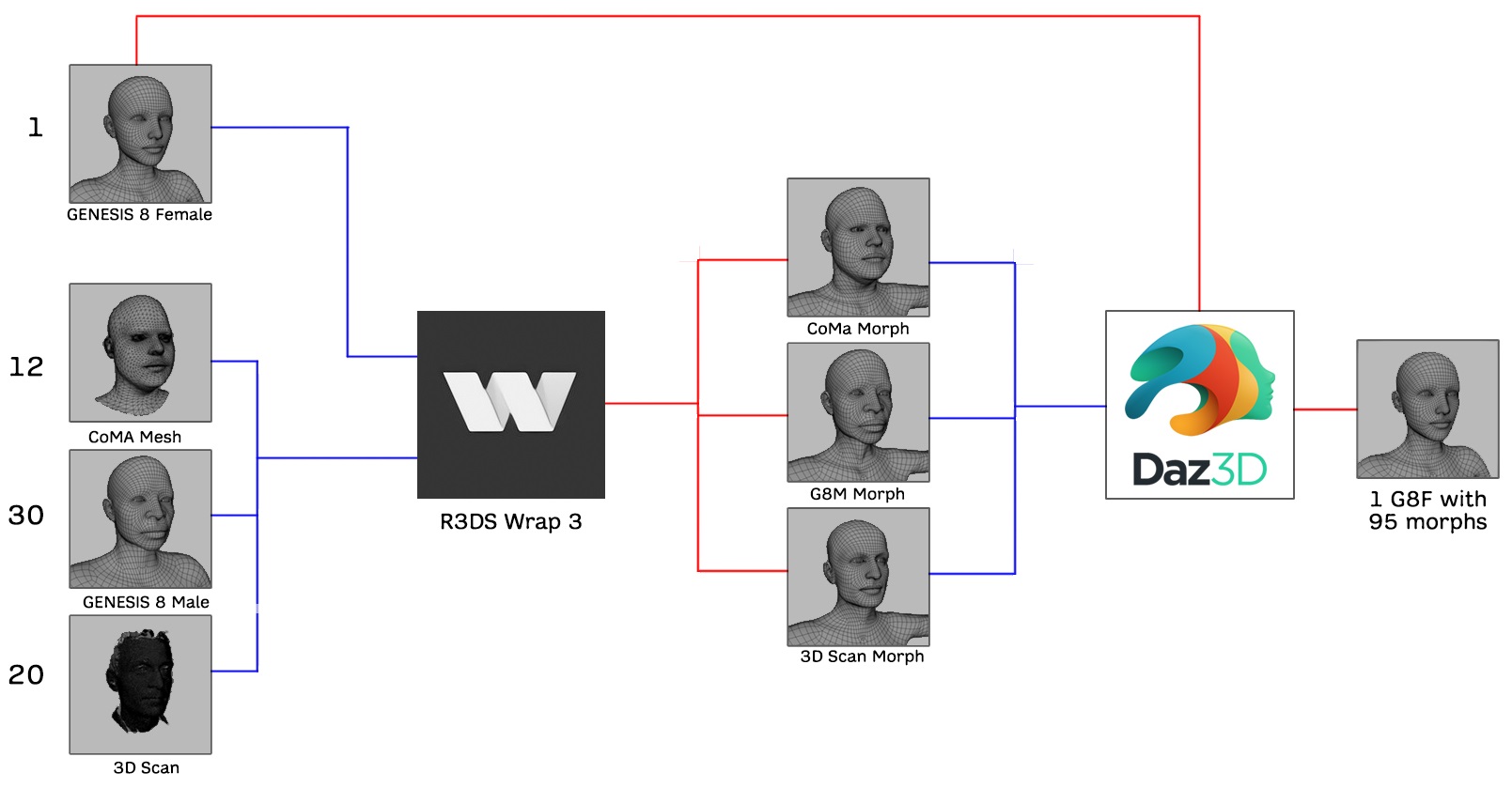}
\caption{The overall data pre-processing pipeline used for generating topologically consistent 4D expression morphs. First, the models/scans from the three sources were mapped to the DAZ Studio’s Genesis~8 Female (G8-F) topology using the Wrap~3 software (note this is not necessary for the 33 female models that are already in the G8-F topology); then, Daz 3D was used to generate these models as morphing from G8-F, thus making the subsequent animation steps easier (\ie, they can be applied to G8-F), while keeping consistent topology across all the models.}
\label{fig:overall}
\end{figure}

\paragraph{Dynamic 3D (4D) datasets}
Given the difficulty of collecting dynamic 3D data, in the literature only a few datasets do exist, which are described in the following.
The Bimghamton University 3D plus time (4D) facial expression dataset~\cite{Yin2008fg} includes 3D facial expression scans captured at 25 frames per second. For each subject, there are six model sequences showing the six prototypical facial expressions (anger, disgust, happiness, fear, sadness, and surprise), respectively. Each expression sequence contains about 100 frames. The database includes 606 3D facial expression sequences captured from 101 subjects, with a total of approximately 60,600 frame models. The resulting database consists of 58 female and 43 male subjects, with a variety of ethnic/racial ancestries, including Asian, Black, Hispanic/Latino, and White. 
The CoMA dataset~\cite{ranjan2018generating} consists of $12$ subjects, each one performing $12$ extreme and asymmetric expressions. Each expression comes as a sequence of fully-registered meshes with $5,023$ vertices. Each sequence is composed of $140$ meshes on average, for a total of $20,466$ scans. 
The D3DFACS dataset~\cite{cosker2011facs} is a collection of dynamic 3D facial expressions, annotated following the Facial Action Coding System. It contains AU sequences from 10 people, with 519 sequences in total. A version of the dataset with scans registered to a known topology (the same of CoMA) is also available~\cite{li2017learning}. 
Finally, we mention VOCASET~\cite{VOCA2019}, a 4D face dataset with about 29 minutes of 4D scans captured at 60fps and synchronized audio from 12 speakers (4 males, and 4 females).

Different from all the above, our dataset is a mixture of real and synthetic identities, and the synthesized expressions include the standard 6 prototypical ones as well as mixed expressions, with 4D sequences including also expression-to-expression transitions, the latter feature making it unique.

\paragraph{Synthetic datasets}
The work of Wood~\etal~\cite{wood2021iccv} proposed to use a procedurally-generated parametric 3D face model in combination with a comprehensive library of hand-crafted assets to render training images with realism and diversity. Machine learning systems for face-related tasks such as landmark localization and face parsing were trained with this data showing that synthetic data can both match real data in accuracy as well as open up new approaches, where manual labeling would be impossible. This work demonstrated it is possible to perform face-related computer vision in the wild using synthetic data alone. 
In particular, in support of our proposed \datasetname{}~4D, authors in~\cite{wood2021iccv} showed that it is possible to synthesize data with minimal domain gap, so that models trained on synthetic data generalize to real in-the-wild datasets.

\section{PROPOSED DATASET}\label{sect:proposed-dataset}
We identified a key missing aspect in the current literature of 4D face analysis, that is the ability of modeling complex, non-standard expressions and transitions between them. Indeed, current models and datasets are limited to the case, where a facial expression is performed assuming a neutral-apex-neutral transition. This does not hold in the real world, where people continuously switch between one facial expression to another. These observations motivated us to generate the proposed \datasetname{}~4D dataset, which is described in the following sections.

\subsection{Source identities}
\datasetname{}~4D includes real and synthetic identities from different sources: \textit{(a)} CoMA identities; \textit{(b)} high-resolution 3D face scans of real identities; \textit{(c)} synthetic identities.

\paragraph{CoMA identities}
The CoMA dataset~\cite{ranjan2018generating} is largely used for the analysis of dynamic facial expressions. An important characteristic of this dataset that contributed to its large use is the fixed topology, according to which all the scans have $5,023$ vertices that are connected in a fixed way to form meshes with 9,976 triangular facets. The dataset includes 12 real identities (5 females and 7 males).

\paragraph{Synthetic identities}
On the Web, a large number of 3D models of synthetic facial characters, either females or males, can be purchased or downloaded for free. Using these online resources, we were able to add 63 synthetic identities (33 females and 30 males) to the data, \textcolor{black}{selecting those that allow editing and redistribution for non-commercial purposes.} \textcolor{black}{Subjects are split in three ethnic groups, Afro (16\%), Asian (13\%), and Caucasian (71\%).} 
Because such identities are synthetic, the resulting meshes are defect free, and perfectly symmetric, which is different from real faces. To make models more realistic, morphing solutions were applied to include face asymmetries.

\paragraph{3D real scans}
We acquired 3D scans of 20 subjects (5 females and 15 males) with a 3DmD HR scanner. Subjects are mainly students and university personnel, 30 years old on average. Meshes have approximately 30k vertices. \textcolor{black}{Written consents were collected for these subjects for using their 3D face scans}.

\subsection{Data pre-processing}
Combining together the identities from the three sources indicated above, we obtained an overall number of 95 identities, 43 females and 52 males. Identities corresponding to synthetic 3D models and 3D scans of real subjects have different topology when compered with CoMA, and a variable number of facets and vertices. Instead, one objective of our dataset was that of providing identities with the same topology as the CoMA dataset (\ie, 5,023 vertices and 9,976 triangular facets).
To this end, we used a workflow that involved the joint use of the DAZ Studio~\cite{daz3d} and R3DS Wrap~3~\cite{wrap3} software to homogenize the correspondence of the identity meshes. All identities were converted into morphs of the DAZ Studio’s Genesis~8 Female (G8-F) base mesh using the Wrap~3 software that allows one mesh to be wrapped over another by selecting corresponding points of the two meshes. The wrapped meshes were then associated with the G8-F mesh as morphs. At the end of the process, we got a G8-F mesh with 95 morphs of different identities. 
After animating the facial expressions and before exporting the sequence of meshes, we restored the animated G8-F to the original topology of the CoMA dataset. 
The overall process is illustrated in Figure~\ref{fig:overall}.

\begin{figure}[!t]
\centering
\includegraphics[width=0.8\linewidth]{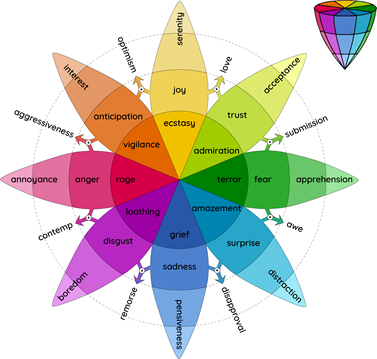}
\caption{Plutchik's wheel of emotions~\cite{plutchik1980}, illustrating expression relations. \label{fig:plutchik}}
\end{figure}

\subsection{Facial expressions}
With the basic Genesis~8 mesh, we also got a set of facial expressions, in the form of morphs that we used for our dataset. The number of presets was expanded by downloading free and paid packages from the DAZ Studio online shop and from other sites.
The base set included 40 different expressions. A paid package of 30 more expressions was added, obtaining a total of 70 different expressions. 
These expressions were classified according to the Plutchik's wheel of emotions~\cite{plutchik1980}, which is illustrated in Figure~\ref{fig:plutchik}. 
\textcolor{black}{Following this organization of expressions, we generated a set of secondary expressions from the eight primary ones (for each primary expression, the number of expression per class is indicated)}: \textit{anger}, AR (6), \textit{fear}, FR (6), \textit{sadness}, SS (13), \textit{disgust}, DT (9), \textit{surprise}, SE (11), \textit{anticipation}, AN (4), \textit{trust}, TT (6), \textit{joy}, JY (15). Details are given in Table~\ref{tab:expr-stat}. 

\begin{table}[!ht]
\centering
\caption{}
\begin{tabular}{lp{0.65\linewidth}}
\hline
\textbf{Primary expression} & \textbf{Expressions} \\
\hline
\textit{Anger}, AR (6) & Angry1, Angry2, Fierce, Glare, Rage, Snarl \\ 
\textit{Fear}, FR (6) & Afraid, Ashamed, Fear, Scream, Terrified, Worried \\ 
\textit{Sadness}, SS (13) & Agony, Bereft, Ill, Mourning, Pain, Pouting, Pouty, Sad1, Sad2, Serious, Tired1, Tired2, Upset \\ 
\textit{Disgust}, DT (9) & Arrogant, Bored, Contempt, Disgust, Displeased, Ignore, Irritated1, Irritated2, Unimpressed \\ 
\textit{Surprise}, SE (11) & Awe, Confused, Ditzy, Drunk1, Frown, Hurt, Incredulous, Moody, Shock, Surprised, Suspicious \\ 
\textit{Anticipation}, AN (4) & Cheeky, Concentrate, Confident, Cool \\ 
\textit{Trust}, TT (6) & Desire, Drunk2, Flirting, Hot, Kissy, Wink \\ 
\textit{Joy}, JY (15) & Amused, Dreamy, Excitement, Happy, Innocent, Laughing, Pleased, Sarcastic, Silly, Smile1, Smile2, Smile3, Smile4, Triumph, Zen \\
\hline
\end{tabular}
\label{tab:expr-stat}
\end{table}

In the dataset, we named the expressions with pairs of names representing the abbreviation of the primary emotion and the facial expression represented, \eg, JY-smile or SE-incredulous. The Genesis~8 mesh also has 70 morphs of facial expressions available, in addition to 95 identity morphs.

\subsection{Creation of expression sequences}
Using the above expression classification, we generated the expression sequences of each identity by iterating through the activation of the expression morphs for each identity morph. 
The dataset includes two types of sequences for each identity: \textit{single expression} and \textit{multiple expressions}.

\paragraph{Single expression}
For each identity, the animation of each morph expression is generated as follows:
\begin{itemize}
\item Frame 0 - neutral expression (morph with weight 0);
\item Random frame between 10 and 50\footnote{With the randomization of the climax frame, we generated a greater variability in the speed of the transition from the neutral to the climax expression and back to the neutral expression for each identity.} - expression climax (morph with weight 1);
\item Frame 60 - neutral expression (morph with weight 0). 
\end{itemize}

\begin{figure}[!t]
\centering
\includegraphics[width=\linewidth]{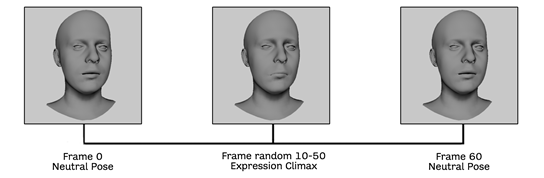} \\
\includegraphics[width=\linewidth]{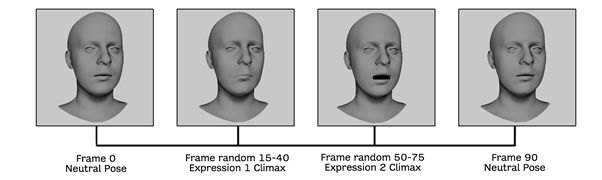}
\caption{Sample frames from a generated sequence: (top) the expression passes from neutral to apex and to neutral again; (bottom) the expression passes from neutral to apex for expr.~1, then to apex for expr.~2, and finally to neutral again.}
\label{fig:n-apex-n}
\end{figure}

\noindent
The meshes in a sequence are named with the name of the expression and the number of the corresponding frame as a suffix (\eg, $Smile\_{01}$. An example is shown in the top row of Figure~\ref{fig:n-apex-n}. 

\paragraph{Multiple expressions}
For each identity, we created mesh sequences of transitions from a neutral expression to a first expression (expr.~1), then from this expression to a second one (expr.~2), then back from the latter to the neutral expression. Also in this case, the climax frames of the two expressions were randomized to obtain greater variability (\ie, the apex frame for each expression can occur at different times of the sequence). Summarizing, these sequences were created following this criterion:
\begin{itemize}
\item Frame 0 - neutral expression (morph expr.~1 weight 0);
\item Random frame between 15 and 40 - morph expr.~1 with weight 1, and morph expr.~2 with weight 0;
\item Random frame between 50 and frame 75 - morph expr.~1 with weight 0, and morph expr.~2 with weight 1;
\item Frame 90 - neutral expression (morph expr.~2 with weight 0).
\end{itemize}

\noindent
Meshes in a sequence are named with the initials of the primary emotions to which the two expressions involved in the animation belong to, followed by the name of the first and second expression plus a numeric suffix for the frame (\eg, $AN$-$AR\_Confident\_Glare\_{01}$.
An example is shown in the bottom row of Figure~\ref{fig:n-apex-n}. 

\subsection{Released data}
Table~\ref{tab:db-summary} reports a quick summary of the main characteristics of the \datasetname{}~4D released data. In particular, we reported the number of identities (male and female), the number of vertices per mesh (same topology for all models), the number of different expressions per identity, the number of sequences that show a neutral-apex expression-neutral transition (6,650 in total); the number of sequences with neutral-expr.~1-expr.~2-neutral transition. Note that, in this latter case, all the possible expression combinations have been generated for a total of 198,550 sequences. 

We also note the neutral-expr-neutral sequences include 60 frames each, with the apex intensity for the expression occurring around frame 30; 90 frames are instead generated for the sequences with an expression-to-expression transition, with the expr.~1 apex and the expr.~2 apex occurring around frame 30 and 60, respectively. 

\begin{table}[!ht]
\centering
\caption{\datasetname{} 4D expression dataset: summary of released data}
\label{tab:db-summary}
\begin{adjustbox}{width=\columnwidth,center}
\begin{tabular}{ccccc}
\hline
\textbf{\#IDs (m/f)} & \textbf{\#vert} & \textbf{\#exprs.} & \textbf{\# n-exp-n}/\# f & \textbf{\# n-exp1-exp2-n}/\# f\\
\hline
95 (52/43) & 5,023 & 70 & 70*95 / 60 & 2090*95 / 90\\
\hline
\end{tabular}
\end{adjustbox}
\end{table}

\begin{figure}[!t]
\centering
\includegraphics[width=\linewidth]{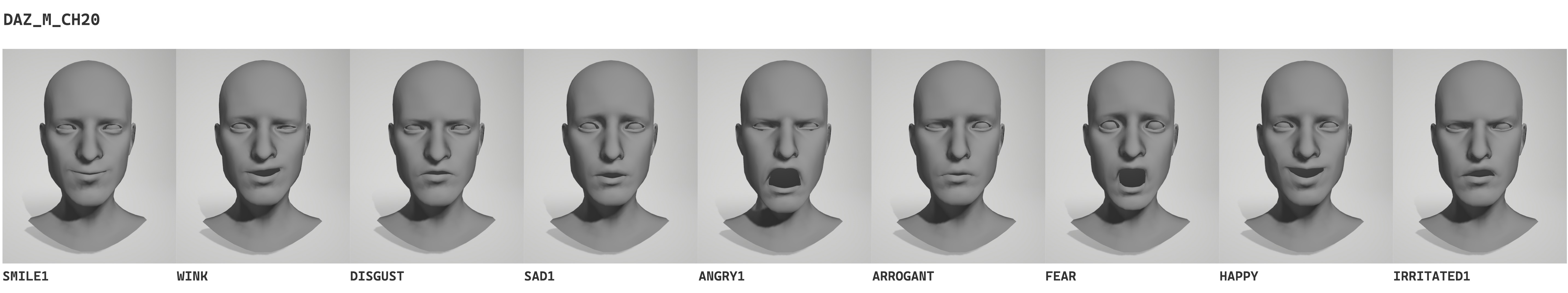}\\
\includegraphics[width=\linewidth]{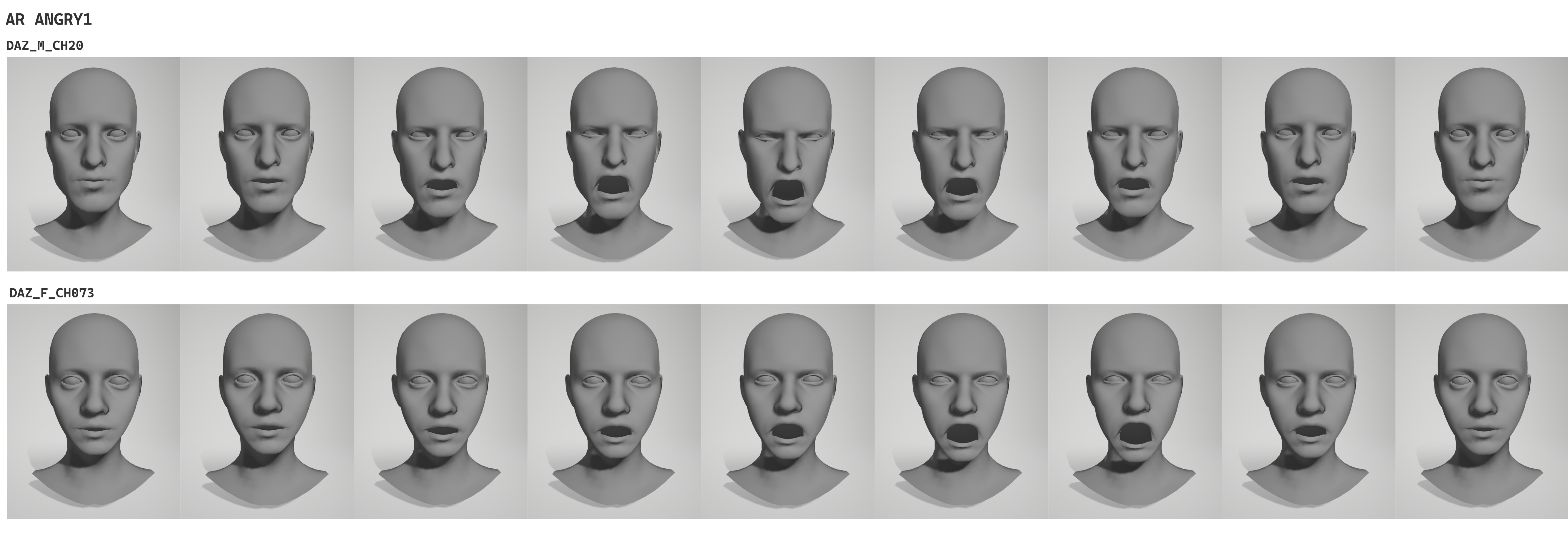}\\
\includegraphics[width=\linewidth]{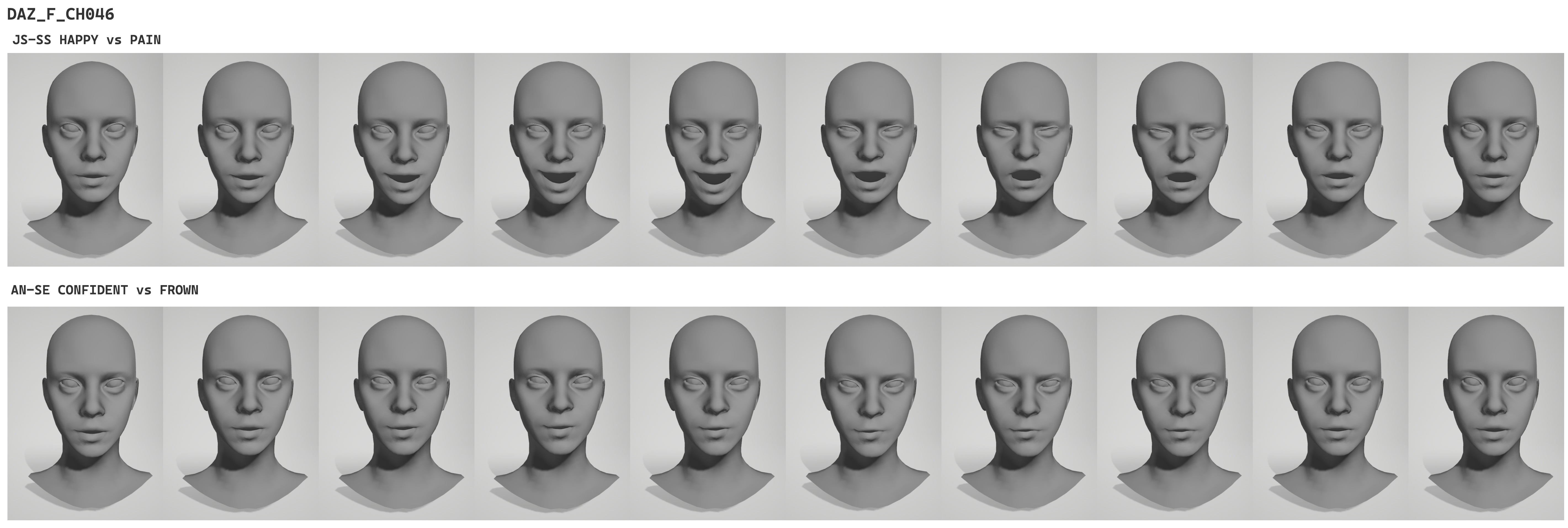}
\caption{Examples frames from generated sequences: (top) apex frames of nine expression sequences for subject $DAZ\_M_CH20$; (middle) \textit{angry} expression for a male ($DAZ\_M\_CH020$) and a female ($DAZ\_F\_CH073$) subject; (bottom) For subject $DAZ\_F\_CH046$ the transitions \textit{happy}-\textit{pain}, and \textit{confident}-\textit{frown} are shown.} 
\label{fig:example-seq}
\end{figure}

Some examples of the generated sequences are illustrated in Figure~\ref{fig:example-seq}. In the top row, the apex frames of nine expression sequences (\ie, smile, wink, disgust, sad, angry, arrogant, fear, happy irritated) of a male synthetic subject are illustrated. 
The second and third row compare frames of an \textit{angry} expression for a male and a female subject. The two bottom rows, instead, show the transitions \textit{happy}-\textit{pain}, and \textit{confident}-\textit{frown} for a given subject.

\section{EXPERIMENTATION}\label{sect:experimentation}
In the following, we report a baseline evaluation for the proposed dataset. We are interested in assessing to what extent our dataset, composed of re-parameterized real scans and totally synthetic sequences, compares to a reference dataset of real scans. We do this by evaluating the task of landmark-based 3D model fitting. As reference datasets to compare with, we chose CoMA and D3DFACS as they share the same mesh topology as \datasetname{}~4D, and are composed of 4D expression sequences. They are also common benchmarks employed in other recent studies~\cite{Bouritsas:ICCV2019, ranjan2018generating}. For a consistent comparison and fulfill our goal, given the way larger amount and variability of sequences included in \datasetname{}~4D, we selected $1,222$ sequences from it, corresponding to the $7$ standard expressions, to make it comparable in size and content to CoMA and D3DFACS.
Following similar previous works~\cite{Bouritsas:ICCV2019, otberdoutcvpr2022}, we performed experiments by splitting the data into train and test. To make sure they do not overlap, in one case, we divided the data based on the identities (Identity Split), in the other, based on expressions (Expression Split). In both the cases, we performed a 4-fold cross validation. 

\subsection{3D Expression fitting}
Since the main focus of \datasetname{}~4D is on expressions, we decided to exclude the problem of identity reconstruction, to avoid ambiguities in the results. The goal is to fit a neutral (not average) 3D face of a subject $\mathbf{S}^n \in \mathbb{R}^{N \times 3}$ to a target expressive face $\mathbf{S}^e$ guided by a set of 3D landmarks $Z^{e} \in \mathbb{R}^{68 \times 3}$. For evaluation, we set up a baseline by first comparing against standard 3DMM-based fitting methods. Similar to previous works~\cite{ferrari2015dictionary, FLAME:SiggraphAsia2017}, we fit $\mathbf{S}^n$ to the set of target landmarks $Z^{e}$ using the 3DMM components. Since the deformation is guided by the landmarks, we first retrieved the landmark coordinates in the neutral face by indexing into the mesh, \ie, $Z^n = \mathbf{S}^n(\mathbf{I}_z)$, where $\mathbf{I}_z \in \mathbb{N}^{68}$ are the indices of the vertices that correspond to the landmarks. We then found the optimal deformation coefficients that minimize the Euclidean error between the target landmarks $Z^{e}$ and the neutral ones $Z^{n}$, and use the coefficients to deform $\mathbf{S}^n$. 
We experimented the standard PCA-based 3DMM and the DL-3DMM~\cite{ferrari2015dictionary}. We also evaluated against recent deep models, including the Neural3DMM~\cite{Bouritsas:ICCV2019} and the very recent S2D-Dec~\cite{otberdoutcvpr2022}. In order to use Neural3DMM as a fitting method, we used the modified architecture as defined in~\cite{otberdoutcvpr2022}, where the model was trained to generate an expressive mesh given its neutral counterpart and the target landmarks $Z^{e}$ as input. The mean per-vertex Euclidean error between the reconstructed meshes and their ground truth was used as measure, as in the majority of works~\cite{Bouritsas:ICCV2019, ferrari2021sparse, PotamiasECCV2020, ranjan2018generating}.

\begin{table*}[!t]
\centering
\caption{Reconstruction error (mm) on expression-independent (left) and identity-independent (right) splits}
\label{tab:comparison-ExprSplit}
\begin{tabular}{l|ccc|ccc}
\hline
& \multicolumn{3}{c}{Expression Split} & \multicolumn{3}{c}{Identity Split}\\
\hline
Method & CoMA & D3DFACS & \datasetname{} 4D & CoMA & D3DFACS & \datasetname{} 4D\\
\hline
PCA & $0.76 \pm 0.73$ &$0.42 \pm 0.44$ & $0.70 \pm 0.81$ & $0.80 \pm 0.73$ & $0.56 \pm 0.56$ & $0.16 \pm 0.17$ \\
\hline
DL3DMM~\cite{ferrari2015dictionary} & $0,86 \pm 0,80$ & $0.73\pm 1.15$ & $0.83\pm 1.03$ & $0.89 \pm 0.79$ & $1.15 \pm 1.50$ &  $0.17\pm 0.18$ \\
\hline
Neural3DMM~\cite{Bouritsas:ICCV2019} & $0.75 \pm 0.85$ & $0.59 \pm 0.86$ & $1.45 \pm 1.43$ & $3.74 \pm 2.34$ & $2.09\pm 1.37$ & $1.41 \pm 1.09$ \\
\hline
S2D-Dec & $0.52 \pm 0.59$ & $0.28\pm 0.31$ & $0.57\pm 1.24$ & $0.55 \pm 0.62$ & $0.27 \pm 0.30$ & $0.10\pm 0.08$ \\
\hline
\end{tabular}
\end{table*}

Table~\ref{tab:comparison-ExprSplit} reports the results. It can be noted that for the expression split, results are similar for all the compared datasets. We argue this represents a piece of evidence that the synthetic expressions are as difficult to reconstruct as the real ones, making them valid to be used in practice. Results for the identity split are instead much lower for the proposed \datasetname{}~4D. Likely, the variability of synthetic identities is lower than that of real ones, being obtained as a result of a generative software process.

\section{DISCUSSION AND CONCLUSIONS}\label{sect:conclusions}
In this paper, a new dataset named \datasetname{}~4D, was presented. Its design and generation was guided by the goal of advancing the research in 4D facial analysis, with a particular focus on dynamic expressions. Compared to current datasets, its unique characteristic is that of including sequences of complex, non-standard expressions. Differently from the existing ones, \datasetname{}~4D also includes dynamic transitions across expressions, extending the standard neutral-peak-neutral setting. All the sequences were generated with randomized velocity for improved realism. The dataset is a combination of real and synthetic identities, while the expressions are fully synthetic. An experimental validation highlights the little domain gap with respect to real expressive scans, making it a valuable resource for real applications.

\section{ACKNOWLEDGMENTS}
This research was partially supported by European Union's Horizon 2020 research and innovation program under grant number 951911 - AI4Media. 
Most of this work was done when Naima Otberdout was at University of Lille. Her work was supported partially by the French National Agency for Research (ANR) under the Investments for the future program with reference ANR-16-IDEX-0004 ULNE and by the ANR project Human4D ANR-19-CE23-0020.

{\small
\bibliographystyle{ieee}
\bibliography{egbib,bibliography,others}
}

\end{document}